\documentclass{article}

\usepackage{arxiv}

\usepackage[utf8]{inputenc} 
\usepackage[T1]{fontenc}    
\usepackage{hyperref}       
\usepackage{url}            
\usepackage{booktabs}       
\usepackage{amsfonts}       
\usepackage{nicefrac}       
\usepackage{microtype}      
\usepackage{lipsum}		
\usepackage{graphicx}
\usepackage{natbib}
\usepackage{doi}

\usepackage{float}
\usepackage{algorithm}
\usepackage{algpseudocode}
\usepackage{amsmath, amssymb} 

\title{SILS: Strategic Influence on Liquidity Stability and Whale Detection in Concentrated-Liquidity DEXs}

\author{
Ali RajabiNekoo \\
Department of Computer Engineering \\
SR.C, Islamic Azad University,  \\
Tehran, Iran \\
\texttt{ali.rajabinekoo@iau.ir} 
\And
Laleh Rasoul \\
Department of Computer Engineering \\
SR.C, Islamic Azad University \\
Tehran, Iran \\
\texttt{laleh.rasoul@iau.ir}
\And
Amirfarhad Farhadi \\
Department of Computer Engineering \\
Iran university of Science and Technology \\
Tehran, Iran \\
\texttt{am\_fahadi@mail.iust.ac.ir}
\And
Azadeh Zamanifar \\
Department of Computer Engineering \\
SR.C, Islamic Azad University\\
Tehran, Iran \\
\texttt{azamanifar@iau.ac.ir}
}

\renewcommand{\shorttitle}{\textit{arXiv} Template}

\begin{document}
\maketitle
\begin{abstract}
Traditional methods for identifying impactful liquidity providers (LPs) in Concentrated Liquidity Market Makers (CLMMs) rely on broad measures, such as nominal capital size or surface-level activity, which often lead to inaccurate risk analysis. The SILS framework offers a significantly more detailed approach, characterizing LPs not just as capital holders but as dynamic systemic agents whose actions directly impact market stability. This represents a fundamental paradigm shift from the static, volume-based analysis to a dynamic, impact-focused understanding. This advanced approach uses on-chain event logs and smart contract execution traces to compute Exponential Time-Weighted Liquidity (ETWL) profiles and apply unsupervised anomaly detection. Most importantly, it defines an LP’s functional importance through the Liquidity Stability Impact Score (LSIS), a counterfactual metric that measures the potential degradation of the market if the LP withdraws. This combined approach provides a more detailed and realistic characterization of an LP’s impact, moving beyond the binary and often misleading classifications used by existing methods. This impact-focused and comprehensive approach enables SILS to accurately identify high-impact LPs—including those missed by traditional methods—and supports essential applications like a protective oracle layer and actionable trader signals, thereby significantly enhancing DeFi ecosystem. The framework provides unprecedented transparency into the underlying liquidity structure and associated risks, effectively reducing the common false positives and uncovering critical false negatives found in traditional models. Therefore, SILS provides an effective mechanism for proactive risk management, transforming how DeFi protocols safeguard their ecosystems against asymmetric liquidity behavior.
\end{abstract}


\section{Introduction}\label{sec:Introduction}
\indent In recent years, the emergence of decentralized exchanges (DEXs) employing a concentrated liquidity architecture—exemplified by Uniswap v3 \cite{uniswapV3Whitepaper}—has deeply transformed the design of blockchain-based financial markets. These platforms, which operate via automated market makers (AMMs), enable liquidity providers (LPs) to concentrate their capital within specific price ranges. While this innovation significantly improves capital efficiency, it also renders the strategic landscape among players more complex and sensitive.\\
\indent Whale Detection in Blockchain Markets Identifying and tracking whales in financial markets—particularly within the blockchain ecosystem—is a significant aspect of market behavior analysis and aligns with the primary focus of this study. Several methods have been proposed for whale detection. One of the most common methods involves analyzing wallet behavior and size, including examining large transactions, deposit and withdrawal frequencies, and interactions with decentralized (DEX) or centralized exchanges. Projects such as \cite{nansenAI} and Arkham \cite{arkm} utilize this approach. Nansen enables monitoring of "smart money" activity by tagging wallets and identifying whales, while Arkham Intelligence leverages artificial intelligence to reveal the identities of some large wallets and visually map the relationships between addresses.\\
\indent Another approach is large transaction monitoring. This method sets a threshold for transaction volume and analyzes counterparties to determine whether the activity involves smart contracts or platforms such as staking protocols or DEXs. Social media analysis and news monitoring also contribute to whale identification. Additionally, institutions such as the U.S. Department of the Treasury can track major capital movements and detect patterns that may be intended to influence the market.\\
\indent A further technique is clustering wallet addresses, where addresses are categorized with labels such as exchange, Decentralized Autonomous Organization (DAO), or whale, or segmented into different behavioral groups using clustering algorithms. Although all blockchain transactions are publicly accessible, transparent, and searchable, the real identities of users are not directly revealed. To protect privacy, users often employ multiple distinct addresses. This inherent design makes address clustering essential for inferring common ownership of multiple addresses and linking them to a single underlying entity. Due to the nature of this process, address clustering cannot provide exact or definitive solutions and therefore relies on approximations and greedy algorithms.\\
\indent Finally, wallet clustering is constrained by the availability of data. Given the pseudonymous nature of blockchain and the ease of creating new wallets, even advanced clustering algorithms cannot ensure perfect accuracy. The permissionless architecture of blockchains inherently limits the certainty of these methods.\\
\indent Within this setting, large‐scale market actors (whales) can produce stabilizing or destabilizing effects on price and liquidity equilibrium through targeted liquidity allocations or high‐volume trades. These phenomena—referred to as the "strategic influence on liquidity stability"—are heavily influenced by the decisions of individual players and the subsequent reactions of other ecosystem participants. Moreover, by leveraging game‐theoretic tools, particularly under conditions of incomplete information and asymmetric players, one can achieve a deeper understanding of market behavior and the mechanisms that safeguard against manipulation. In this study, we propose a conceptually grounded, implementable framework in which LPs, traders, and whales serve as strategic players interacting within concentrated‐liquidity pools. \\
\indent Finally, by introducing an approach for whale detection and analyzing optimal counterstrategies, our framework, which focuses on asymmetric‐player analysis in concentrated‐liquidity AMMs can inform the design of anti‐manipulation policies at the protocol level and generate actionable trading signals for market participants. The aim of this research is to provide a theoretical and practical foundation for understanding and enhancing liquidity stability in modern DEXs.\\
\indent We first establish a game‐theoretic framework based on a concentrated‐liquidity DEX, from which we extract transaction records and event logs directly from the blockchain and the pool’s smart contract. Next, we apply clustering techniques to categorize liquidity providers according to two key metrics Time Weighted Liquidity (TWL) and average price impact (PI). Next, we apply anomaly detection algorithms to identify the cluster of actors whose behavior significantly deviates from that of the general LP population. These anomalies correspond to whales, because their disproportionately large liquidity stakes and significant market impact distinguish them from ordinary participants. Finally, for each detected anomaly, we compute a "Degradation" metric—quantifying the decline in liquidity stability that would result from the whale’s removal. By integrating clustering, anomaly detection, and Degradation analysis, our approach automatically and accurately discriminates whales from non‑whale liquidity providers.\\
\indent As part of our protocol-level safeguards against manipulation and in order to maintain liquidity stability in concentrated-liquidity DEXs, we redesign the LP withdrawal (burn) workflow by introducing a protective oracle layer that intercepts burn requests before they reach the liquidity pool’s smart contract. This oracle conducts a real-time stability assessment using key equilibrium metrics—including price volatility, market depth, and the Liquidity Stability Impact Score (LSIS)—to evaluate whether the requested withdrawal may trigger disruptive market imbalances. Only requests that meet the defined stability criteria are authorized and executed; otherwise, they are suspended or rejected to ensure systemic resilience.\\
\indent This study builds upon a series of key prior works. Uniswap V3 \cite{uniswapV3Whitepaper}, as the first implementation of concentrated liquidity architecture, provides the technical foundation for examining liquidity resilience. In the domain of whale identification, \cite{herremans2022forecasting} have mapped critical paths for analyzing large actors’ behavior by leveraging social and on-chain data, while \cite{zhang2023out}, through the ExpLTV model, have advanced whale detection in mobile gaming. Regarding anomaly detection in blockchain, \cite{han2024mt} have played a critical role by employing the $MT^2AD$  model and graph neural networks, and \cite{li2024stateguard} have contributed through StateGuard, a framework for detecting security vulnerabilities in DEX smart contracts. Additionally, \cite{mo2023toward} proposed VeriOracle, an oracle-centric approach to reduce price feed attacks in blockchain.\\
\indent Building on these contributions, our research aims to fill existing gaps in analyzing liquidity providers’ behavior and designing an active protective layer in DEXs. It offers a novel approach to analyzing and enhancing liquidity resilience in decentralized exchanges based on concentrated liquidity.\\
\indent The remainder of this paper is organized as follows. Section~\ref{sec:Background} provides essential background information necessary for understanding the context of our study within decentralized finance. Section~\ref{sec:RelatedWorks} offers a comprehensive review of related works concerning liquidity analysis and influential actor identification in DeFi. Following this, Section~\ref{sec:ProblemDefinition} formally defines the problem addressed by this research, including the foundational mathematical concepts of Uniswap V3. Our proposed SILS framework, which includes a comprehensive methodology for data collection, liquidity state tracking, top LP identification, and counterfactual analysis, is thoroughly explained in Section~\ref{sec:methodology}. Section~\ref{sec:ResultsDiscussion} presents the results and discussion of our experimental findings. Finally, Section~\ref{sec:LimitationsFutureWork} outlines the limitations and future work, and Section~\ref{sec:Conclusion} concludes the paper.\\
Our contribution is:
\begin{itemize}
  \item \textbf{A Novel Game‑Theoretic Framework for Concentrated‑Liquidity DEXs:} \\
    \indent We present a unified and implementable framework that models interactions between liquidity providers (LPs), traders, and whales in concentrated-liquidity pools, capturing their strategic behaviors despite incomplete information and player asymmetry.
  \item \textbf{Integrated Clustering‑Based Whale Detection:}\\
    \indent We use unsupervised clustering on key on‑chain metrics Time Weighted Liquidity (TWL) and average Price Impact (PI)—to group LPs with similar behaviors, then apply anomaly detection algorithms to automatically identify whales as statistical outliers.
  \item \textbf{Liquidity Stability Impact Score (LSIS) for Counterfactual Analysis:}\\
    \indent We propose and formalize a novel metric, the Liquidity Stability Impact Score (LSIS), to quantify the destabilizing effect on market liquidity caused by the hypothetical removal of a specific liquidity provider. LSIS integrates key market stability indicators such as price volatility, pool depth, and average price impact, offering a rigorous and interpretable measure for counterfactual sensitivity analysis.
  \item \textbf{Exponential Time-Weighted Liquidity (ETWL):}\\
    \indent We introduce a novel use of Exponential Time Weighted Liquidity (ETWL) metrics to characterize the behavioral patterns of liquidity providers over time. By weighting liquidity contributions based on their duration and consistency, we are able to distinguish between transient, opportunistic actors and persistent, high-impact providers—often corresponding to whales. This temporal profiling enhances the accuracy of anomaly detection and clustering, and provides a robust foundation for assessing systemic risks associated with sudden liquidity withdrawals.
  \item \textbf{Oracle‑Based Protective Gatekeeping Mechanism:}\\
    \indent We introduce and prototype a protective oracle module that intercepts LP burn requests, evaluates critical liquidity‑stability metrics off‑chain, and conditionally permits or blocks withdrawals at the protocol level to prevent market manipulation and preserve DEX equilibrium.
\end{itemize}

\section{Background}\label{sec:Background}
\subsection{Decentralized Exchanges and Automated Market Makers (DEXs \& AMMs)} 
\indent Decentralized exchanges (DEXs) are a cornerstone of decentralized finance (DeFi). By leveraging innovative architectures, DEXs enable the peer-to-peer exchange of digital assets without intermediaries. The architectures used in DEXs are primarily categorized into two models: the order book model and the automated market maker (AMM) model. While the order book model is similar to the structure of traditional centralized exchanges by relying on matching buy and sell orders between users, the AMM model has fundamentally transformed the structure of decentralized trading by removing this dependency.\\
\indent In the AMM model, trades are executed not through direct order matching but through interactions with liquidity pools. Liquidity providers (LPs) deposit pairs of assets (such as ETH and USDT) into a shared pool, allowing traders to execute swaps directly with the pool itself. In other words, the counterparty to each trade is the pool. For each transaction, a trading fee is charged to the user and distributed according to among LPs based on their share in the pool.\\
\indent Classic AMM models, such as Uniswap V2, use a simple mathematical pricing relationship where the product of the quantities of the two tokens in the pool remains constant ($x \cdot y = k$). This mechanism ensures that the price of one token adjusts automatically and immediately when there is an inflow or outflow of the other token. Although this approach simplifies trade execution and eliminates the need for user coordination, it can lead to challenges such as slippage and capital inefficiency. The fundamental difference between AMMs and order books lies in price discovery and trade execution mechanisms. In the order book model, active participation from users to place and match orders is necessary, and trades occur only if there is a corresponding counter-order. In contrast, in the AMM model, price discovery and trade execution occur instantly based on the pool’s asset balances. This approach enables DEXs to facilitate transactions in a permissionless, censorship-resistant, and fully decentralized environment (Uniswap \cite{uniswapV3Whitepaper}).

\subsection{Concentrated Liquidity} 
\indent One of the key innovations introduced in the third generation of the Uniswap protocol was the concept of concentrated liquidity, which fundamentally transformed the traditional architecture of AMMs. In earlier models such as Uniswap v2, liquidity providers (LPs) were required to distribute their assets across the entire possible price range $[0, \infty)$. This means that regardless of the current trading price, liquidity was uniformly spread across all price intervals. As a result, a significant part of LPs’ capital remained idle at price points where no trades were happening, generating no fees and leading to capital inefficiency.\\
\indent In contrast, Uniswap v3, with its concentrated liquidity design, allows LPs to specify a custom price range within which they are willing to provide liquidity. Each liquidity position is defined by two boundary prices, shows that the LP is only providing liquidity within that specified price interval. This feature enables a smaller amount of capital to generate higher returns in high-activity price ranges, significantly increasing the effective liquidity at the market equilibrium point. Therefore, Uniswap v3 achieves a significant improvement in capital efficiency compared to its previous versions (Uniswap \cite{uniswapV3Whitepaper}).

\subsection{Counterfactual Analysis} 
\indent Counterfactual analysis is a central pillar of modern causal inference, focusing on evaluating "what-if" scenarios. This approach, extensively developed by \cite{pearl2009counterfactual}, allows us to move beyond observational data and analyze outcomes that could have occurred under different conditions than what actually occurred. Unlike traditional statistical analyses, which primarily emphasize correlations, counterfactual analysis aims to uncover and quantify causal relationships between variables, examining how the final outcome would have changed if one or more variables had taken different values in the past.\\
\indent The significance of counterfactual analysis lies in its ability to provide a deeper understanding of the underlying mechanisms of a system. By simulating various scenarios, this approach enables researchers to assess the potential consequences of different decisions before implementation, thereby playing a critical role in data-driven policy design (\cite{pearl2009counterfactual}).\\
\indent Moreover, counterfactual analysis offers a strict mathematical framework for addressing causal questions, supporting evidence-based decision-making. In an era with large amounts of available data, distinguishing between correlation and causation is essential to avoid misleading conclusions and to enable optimal decision-making. This message is so central that Pearl and Mackenzie authored The Book of Why (\cite{pearl2018why}) to emphasize the importance of this distinction to a broader audience. As detailed in Pearl’s Causality (\cite{pearl2009counterfactual}), counterfactual analysis serves as a powerful tool for achieving this goal and advancing the development of AI systems capable of causal reasoning (\cite{pearl2018why}).

\subsection{Blockchain Oracles} 
\indent Smart contracts (\cite{ethereumWhitepaper}), which form the core of decentralized applications (dApps) and decentralized finance (DeFi), are inherently executed in an isolated and deterministic environment on the blockchain. While this characteristic ensures transparency and verifiability, it also introduces a significant limitation: smart contracts cannot directly access off-chain data, nor can they interact with web APIs, sensor data, or other external sources. This challenge, known as the "oracle problem", is a fundamental issue for blockchain in real-world applications, as many practical functions (such as collateralization, insurance, and event prediction) require real-world data.\\
\indent To address this limitation, oracles have become as a critical blockchain infrastructure, acting as critical intermediaries that relay data from the external world to smart contracts. Structurally, oracles can be categorized into two main types: centralized and decentralized (\cite{al2020trustworthy, pierro2023analysis}). Centralized oracles rely on a single source or node to provide data, a method that may be desirable in some cases due to its simplicity and speed. However, this approach introduces serious risks, including a single point of failure and vulnerability to data manipulation.\\
\indent In contrast, decentralized oracles leverage networks of independent nodes to collect, validate, and transmit data from multiple sources to the blockchain using consensus mechanisms. This architecture not only reduces the probability of errors or data corruption but also provides greater resistance to malicious attacks and systemic disruptions (\cite{al2020trustworthy}). A significant example of decentralized oracle implementation is Chainlink, which securely and tamper-resistently supplies smart contracts with data such as cryptocurrency prices, exchange rates, insurance information, and stock market data. By leveraging external data feeds and multi-source validation, these oracles have significantly reduced  trust-related challenges.\\
\indent On the other hand, on-chain oracles, such as those implemented in Uniswap v3, directly extract the required data from the internal behavior of the blockchain itself. In other words, rather than relying on external sources, these oracles utilize data generated by the network over time. In Uniswap v3, such data includes indicators like the Time Weighted Average Price (TWAP) and Cumulative Log Price (CLP), which are continuously stored within pool contracts in an optimized structure , such as circular arrays (Uniswap \cite{uniswapV3Whitepaper}). This design enables smart contracts to access accurate, stable, and verifiable data without requiring interaction with external oracles. The primary advantage of these oracles lies in their high reliability, which stems from their exclusive reliance on on-chain data. However, they have a notable limitation in their scope of application, as they can only provide information derived from the protocol or network itself and cannot deliver external data, such as asset prices from other markets or real-world event outcomes.\\
\indent Therefore, the role of oracles extends beyond merely "data transmitters", positioning them as protective gatekeepers that support the integrity and security of smart contracts. As noted by  \cite{al2020trustworthy}, the performance quality of a smart contract is directly tied to the accuracy and validity of the data it receives from oracles. Additionally, studies by \cite{ezzat2022blockchain} indicate that oracles not only play a key role in bridging the off-chain and on-chain worlds but can also serve as effective tools for cross-chain interoperability and facilitating interactions among dApps.

\section{Related Works}\label{sec:RelatedWorks}
\indent In this section, we review the research areas related to our study.
\subsection{Anomaly and Whale Detection in Financial Markets and Blockchain} 
The study of whales (entities that either hold substantial assets or have significant influence over markets) has been a recurring theme in various financial domains, including traditional markets and, more recently, the field of cryptocurrencies. In the context of digital assets, research has explored the relationship between large transactions and market dynamics. For instance, \cite{herremans2022forecasting} introduced the Synthesizer Transformer model to predict Bitcoin price volatility jumps. Their model integrates data from the Whale Alert account on X (formerly known as Twitter) alongside on-chain metrics provided by CryptoQuant, a platform specializing in the analysis of blockchain-based financial flows and transaction patterns to interpret market behavior. Their findings highlight the critical role that whale transactions play in driving extreme price fluctuations and contributing to market instability.\\
\indent Our research similarly examines the strategic impact of whales, specifically on liquidity stability within DEXs that utilize concentrated liquidity models. However, whereas  \cite{herremans2022forecasting} focuses solely on predicting volatility, it is worth noting that their analysis relies exclusively on tweets publicly posted by Whale Alert, without any direct monitoring of whale behavior, which inherently constrains the study’s comprehensiveness. Not all decisions made by major market players are publicly disclosed, and the available data may lack complete accuracy or transparency. However, our research goes a step further. In addition to directly identifying high-impact liquidity providers (LPs), we quantify the systemic risk they pose using a metric called the Liquidity Stability Impact Score (LSIS). This enables the implementation of a protective mechanism based on LP impact, which can be integrated into a protocol-level protective oracle.\\
\indent Beyond financial markets, the concept of identifying disproportionate or influential actors has also emerged in other domains. For example, \cite{zhang2023out} proposed the ExpLTV framework, a multi-task learning model for predicting Customer Lifetime Value (LTV) and identifying "game whales" in mobile gaming ecosystems. Their study demonstrates that the specific purchasing behavior of these users can distort traditional LTV prediction models, and it utilizes a deep neural network to distinguish between high- and low-spending players. Although the application domain (mobile gaming) and primary objective (LTV prediction) of this study differ from our focus on DeFi market stability, the core challenge—detecting and modeling the behavior of impactful agents using clustering and anomaly detection techniques—is conceptually similar. Our approach adapts these techniques specifically to the mechanics of concentrated liquidity, with a focus on quantifying the financial implications of LP behavior for exchange stability. It is worth noting that while this paper presents an effective multitask framework for LTV prediction and game whale detection, the study is limited to three industrial datasets and does not explore the model’s generalizability across other game genres or markets. Moreover, leveraging graph-based structures to enhance whale behavior detection and examining user behavior changes over time remain open challenges.\\
\indent In the domain of financial forecasting, deep learning models have been proposed to capture market dynamics. \cite{farhadi2025hybrid} proposed an LSTM-GRU-based model for stock price prediction through technical analysis. However, their approach falls short in modeling complex market dynamics and external factors, limiting its effectiveness in real-world financial forecasting.\\
\indent Anomaly detection has also been reviewed in related studies, as it plays a crucial role in maintaining the security and integrity of blockchain networks by detecting irregular or malicious activities. \cite{lataran2024developing} introduced an anomaly detection framework for fraud identification by integrating a BERT-based neural network with blockchain to detect double-spending behaviors. However, while effective in identifying suspicious activities, the model lacks interpretability, and relying solely on deep learning may be insufficient in the financial domain, where the complexity of market behavior often requires explanation and transparent approaches. Existing approaches often rely on traditional machine learning techniques coupled with feature engineering or graph representation learning methods to analyze network transactions.\\
\indent \cite{han2024mt} proposed the $MT^2AD$ model — Multi-Layer Temporal Transaction Anomaly Detection — for detecting temporal anomalies within the Ethereum network. Their approach leverages Graph Neural Networks (GNNs), modeling temporal information as sequences of network snapshots by integrating transaction timing, directional flow, and inter-token trading patterns. By aggregating multiple crypto transactions, they construct a unified graph, framing anomaly detection as a graph classification task. This paper analyzes the Bitcoin transaction network; however, as noted by the authors, the exploration of unsupervised graph neural network methods and the address labeling problem as a key challenge was not addressed in this study.\\
\indent  Our research similarly employs anomaly detection techniques to identify high-impact Liquidity Providers (LPs) within concentrated liquidity pools. We extract transaction and event log data directly from the blockchain and apply clustering methods based on Exponential Time Weighted Liquidity (ETWL) and average price impact to categorize LPs. Using anomaly detection algorithms, we identify behavioral patterns that significantly deviate from the norm and may indicate whale-like entities.\\
\indent  A key innovation of our approach lies in the novel application of TWL-based behavioral profiling, which enables us to distinguish between transient and persistent high-impact LPs. Furthermore, our framework integrates this detection mechanism into a real-time oracle layer that dynamically evaluates and filters LP withdrawal requests using metrics such as the Liquidity Stability Impact Score (LSIS), price volatility, and market depth—capabilities that are significantly absent in conventional anomaly detection systems.\\
\indent Furthermore, ensuring data security is a fundamental concern in distributed and sensitive environments. Recent studies have demonstrated how blockchain technology can enhance security by providing tamper-proof and decentralized data management. For example, \cite{feng2025ai} applied blockchain to secure data in smart home wireless sensor networks, recording every sensor reading on an immutable ledger to prevent unauthorized access. This work highlights the potential of blockchain as a reliable security foundation, which aligns with our focus on protecting DeFi protocols from malicious actions and ensuring trustworthy operations.\\
\indent  In addition to the topics mentioned, examining the security and correctness of smart contract execution is especially critical in DEXs, where complex multi-contract interactions occur. Design or implementation flaws in smart contracts can lead to serious vulnerabilities. \cite{li2024stateguard} present the first systematic study of state derailment defects in DEX smart contracts. Such defects can cause incorrect, incomplete, or unauthorized changes to the system state, posing significant security risks. They propose a deep learning–based framework called StateGuard to detect such defects. StateGuard constructs an abstract syntax tree (AST) from the smart contract code and leverages graph convolutional networks (GCNs) to analyze features for defect identification. This approach provides a powerful method to detect fundamental flaws in smart contract code, potentially preventing many vulnerabilities at their roots. It contributes to enhancing the overall reliability and security of DEX protocols during the development phase.\\
\indent However, the study faces challenges related to time complexity caused by multiple AST traversals and managing large, interconnected graphs when analyzing DApp contracts. Our approach introduces an oracle-based protective gatekeeping system that, rather than directly intervening in contracts to prevent whale liquidity withdrawals, establishes a protective layer to reduce disruptive and volatile impacts on DEXs.\\
\indent In this context, \cite{mo2023toward} propose VeriOracle, a formal verification framework for automatic detection of unexpected price feeds in smart contracts. By creating a formal semantic model of price oracles on the blockchain, VeriOracle monitors smart contract states and identifies anomalous price feed transactions in real time. This method effectively counters oracle price manipulation attacks and can prevent major financial exploits. Notably, this paper directly addresses security issues in price oracles, a critical vulnerability in DeFi. VeriOracle employs formal verification rather than machine learning to detect attack patterns, offering a robust approach to identifying and mitigating price input problems and thus preventing large-scale financial attacks.\\
\indent Centralized liquidity DEXs, due to their AMM nature, perform accurate price calculations, as long as liquidity itself does not induce abnormal price volatility. For this reason, our SILS method aims to secure liquidity by requiring LPs to withdraw controlled and validated amounts. This prevents excessive liquidity withdrawals from causing volatile price swings, helping maintain reasonable prices. However, it is important to note that the study by \cite{mo2023toward} does not comprehensively evaluate the system’s performance at scale across various DeFi protocols, including Uniswap, SushiSwap, and others. Moreover, VeriOracle is based on formal verification and does not use machine learning to identify attack patterns.\\
\indent Previous research has addressed various security aspects of DEXs, including analyzing smart contract coding flaws and the evaluation of oracle-provided data. However, the aim of this review is to emphasize the need for an active, real-time protective layer at the protocol level that can detect and mitigate suspicious or destabilizing behaviors before attacks occur. Accordingly, our proposed framework utilizes a native, real-time oracle that tracks LP withdrawal requests, analyzes the stability status of the liquidity pool, and implements countermeasures. This mechanism functions as an anti-manipulation policy, preventing malicious behaviors during critical moments. Such an approach complements prior security measures and offers an innovative solution to reduce risks in Concentrated Liquidity Market Maker (CLMM) protocols.\\
\indent Furthermore, a review of existing studies shows that although the impact of whales on cryptocurrency volatility and anomaly detection in blockchain transactions have been explored, no comprehensive protocol-level framework has yet been developed specifically to identify high-impact LPs and mitigate systemic risks in DEXs with concentrated liquidity.
The present study addresses these gaps by introducing several fundamental innovations, as outlined in the Introduction section.

\subsection{Event Retrieval and Gas Optimization in Smart Contracts} 
\indent One of the most important and challenging issues in today's era is reducing the gas costs associated with data storage in decentralized applications (DApps). As the volume of data that must be stored in smart contracts increases, gas costs rise significantly (\cite{ethereumGasDocs}). To reduce these costs, effective strategies to minimize the amount of data stored directly on smart contracts are essential (\cite{solidityABISpec,ethereumSmartContracts}). Among these approaches is finding alternative methods for storing information off-chain, without the need to maintain it directly in the smart contract’s memory.\\
\indent One such approach, proposed by \cite{kostamis2024data}, involves leveraging event logs (\cite{etherscanEventLogs}) and retrieving logs generated by smart contracts. This technique has become a widely adopted practice among DApp developers, who add events as payloads within transactions to minimize gas consumption (\cite{li2023understanding}). However, a fundamental challenge remains: efficiently accessing and retrieving event data and logs from the Ethereum network. Typically, these applications scan only the events associated with their specific smart contracts and use the JSON-RPC API (\cite{ethereumJsonRpcDocs,ethereumNodesClientsDocs}) over HTTP or WebSocket connections to receive updates on new contract events, ensuring they remain informed of the contract’s latest state.\\
\indent In another study, \cite{morichetta2025event} utilized event extraction techniques and the Extensible Event Stream (XES) format (\cite{wynn2024ieee}) to capture and store blockchain data. Their goal was process mining, with output containing a list of extracted events, including function names, inputs/outputs, and other parameters formatted in XES. This approach is notable because it leverages the smart contract’s Application Binary Interface (ABI) (\cite{solidityABISpec}) to decode inputs and derive meaningful insights about user interactions. Additionally, event arguments stored as topics can be decoded via the ABI. Although this study was not primarily focused on comprehensive data retrieval, incorporating these arguments into XES files enables full data extraction. However, this method is not efficient at a large scale, as it may struggle in decentralized applications like Uniswap, which rely on multiple smart contracts and distributed event logs. It cannot perform targeted extraction when logs are scattered across different contracts and blocks. \\
\indent Motivated by the need to reduce gas costs and enhance data processing efficiency in decentralized finance (DeFi) applications like Uniswap, the present study adopts a targeted event log extraction approach. A key component in designing behavioral indicators for liquidity providers (LPs) and identifying whales is comprehensive analysis of historical user interactions across multiple smart contracts on a global scale. This objective is achievable by processing network-wide registered logs. Furthermore, this approach not only reduces gas consumption but also facilitates richer data availability for advanced analysis.\\

\section{Problem Definition}\label{sec:ProblemDefinition}
\indent In decentralized financial (DeFi) markets, actors who are able to executing large transactions can significantly influence price volatility, liquidity depth, and the overall health of the ecosystem. Identifying these entities, often referred to as "whales", typically relies on static, volume-based metrics, such as the total value locked (TVL) of an address or the percentage of a token’s circulating supply that they hold.\\
\indent However, these static definitions fail to distinguish "strategic significance" from mere "nominal size". An actor may passively hold a large volume of an asset without actively contributing to market dynamics, while another actor with a smaller holding may play a critical role in providing liquidity within key price ranges, making them essential for market stability. Consequently, the existing literature faces a critical gap: the lack of a functional definition of "whale" that captures the actor’s real role in system dynamics and stability.\\
\indent This study seeks to address these challenges by posing the following research questions:
\begin{itemize}
    \item \textbf{RQ1:} How can we develop a functional definition of a "whale" that identifies the role of a LP in maintaining the stability of a DEX?
    \item \textbf{RQ2:} How can we quantitatively measure the "systemic importance" of an actor beyond traditional asset volume metrics?
\end{itemize}
\indent To answer these questions, this paper proposes an innovative framework called SILS (Strategic Influence on Liquidity Stability), which draws upon game theory and complex systems analysis to introduce a new perspective. Within this framework, a "whale" is defined and identified not by the volume of assets held, but by their critical impact on liquidity stability, measured through price impact metrics. This impact is evaluated through a counterfactual analysis method, comparing system stability under hypothetical scenarios both with and without the actor in question. This approach enables the identification of actors whose removal would cause the greatest degradation in the stable functioning of the market.\\
\subsection{Mathematical Foundations and Notation}
\indent To fully understand the proposed SILS methodology, it's essential to first review the core mathematical concepts underlying concentrated liquidity pools in Uniswap V3. Our model is directly built upon these principles to assess liquidity stability. The following section introduces the key notations and formulas used throughout this study.\\
\begin{table}[ht]
    \centering
    \caption{Mathematical Symbols and Definitions for Uniswap V3}
    \begin{tabular}{llp{7cm}}
        \hline
        \textbf{Symbol} & \textbf{Name} & \textbf{Description} \\
        \hline
        $P$ & Price & Relative price of tokens (\texttt{token0} in terms of \texttt{token1}). \\
        $i$ & Tick Index & Integer representing price in logarithmic scale. \\
        $\sqrt{P}$ & Square Root Price & Primary storage format for optimized calculations. \\
        \texttt{sqrtPriceX96} & Fixed-Precision Price & Price as $\sqrt{P} \times 2^{96}$ in fixed-point precision. \\
        $L$ & Liquidity & Virtual liquidity amount in a price range. \\
        $\Delta x, \Delta y$ & Swapped Amounts & Swapped amounts of \texttt{token0} and \texttt{token1}. \\
        $i_l, i_u$ & Lower and Upper Ticks & Lower and upper ticks of a liquidity position (\texttt{tickLower}, \texttt{tickUpper}). \\
        $L_{\text{active}}$ & Active Liquidity & Total liquidity for positions containing current tick. \\
        \hline
    \end{tabular}
    \label{tab:uniswap_symbols}
\end{table}
\subsection{Key Formulas of Uniswap V3}
\subsubsection{\textbf{Price and Tick Relationship:}}
In Uniswap V3, prices are represented discretely using ticks. The relationship between the price $P$ and the tick index $i$ is defined as follows:
\[
P(i) = 1.0001^{i}
\]
This logarithmic relationship implies that each consecutive tick represents a constant price change of 0.01\%.
\subsubsection{\textbf{Price Formula and \texttt{sqrtPriceX96}:}}
To avoid rounding errors and reduce gas costs in smart contract computations, Uniswap uses the square root of price, denoted as $\sqrt{P}$, as the conventional price representation. This value is stored using fixed-point math with a Q96 format:
\[
\sqrt{P} = \frac{\texttt{sqrtPriceX96}}{2^{96}}
\]
\subsubsection{\textbf{Liquidity and Swap Calculation:}}
In a price range where liquidity ($L$) remains constant, Uniswap V3 operates like a constant product market maker ($x \cdot y = k$), where swap calculations are based on changes in $\sqrt{P}$. If the price shifts from $\sqrt{P_a}$ to $\sqrt{P_b}$, the swap amounts ($\Delta x$ and $\Delta y$) can be derived from the following formulas:
\[
\Delta x = L \cdot \left( \frac{1}{\sqrt{P_b}} - \frac{1}{\sqrt{P_a}} \right)
\]
\[
\Delta y = L \cdot \left( \sqrt{P_b} - \sqrt{P_a} \right)
\]
These equations form the basis for calculating price impact. In this model, for a given swap with known volume, a lower active liquidity ($L$) leads to a larger price movement ($\sqrt{P_b} - \sqrt{P_a}$), and thus a higher impact.
\subsubsection{\textbf{Active Liquidity ($L_{\text{active}}$):}}
Active liquidity at a given tick (i.e., the current price range) is defined as the net change in liquidity ($\Delta L$) across all initialized ticks up to that tick. This value, denoted as $L_{\text{active}}$, is used to assess pool depth and to compute the price impact during swaps.
\subsection{Core Formulas}
All these formulas are taken from the Uniswap \cite{uniswapV3Whitepaper} whitepaper.
\subsubsection{\textbf{Tick-to-Price Relationship:}}
This formula  defines the price P as an exponential function of the tick index i:
$$P(i) = 1.0001^i$$
\subsubsection{\textbf{Conversion from \texttt{sqrtPriceX96} to Price:}}
This equation calculates the amount of token0 swapped, denoted by \(\Delta x\), based on the available liquidity \(L\) and the change in price from \(\sqrt{P_a}\) to \(\sqrt{P_b}\):
$$\sqrt{P} = \frac{\text{sqrtPriceX96}}{2^{96}}$$
\subsubsection{\textbf{Swap Volume Formula for \texttt{token0} ($\Delta x$):}}
This equation calculates the amount of token0 swapped, denoted by \(\Delta x\), based on the available liquidity \(L\) and the change in price from \(P_a\) to \(P_b\):
$$\Delta x = L \cdot \left| \frac{1}{\sqrt{P_b}} - \frac{1}{\sqrt{P_a}} \right|$$
\subsubsection{\textbf{Swap Volume Formula for \texttt{token1} ($\Delta y$):}}
This formula computes the amount of token1 swapped, $\Delta y$, based on liquidity L and the change in price from \(\sqrt{P_a}\) to \(\sqrt{P_b}\):
$$\Delta y = L \cdot \left| \sqrt{P_b} - \sqrt{P_a} \right|$$

\section{Methodology}\label{sec:methodology}
\begin{figure}[H]
    \centering
    \includegraphics[width=0.85\textwidth]{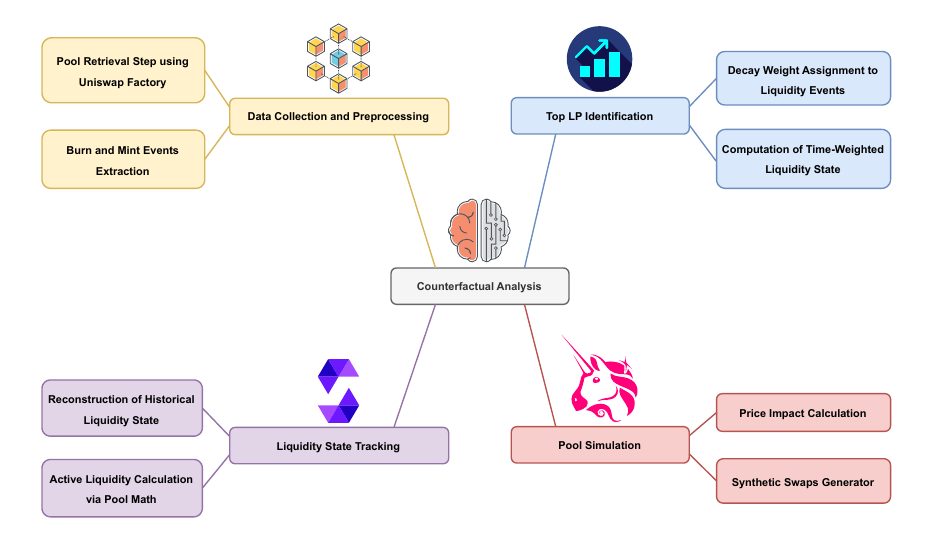}
    \caption{Graphical Abstract: Workflow of whale analysis on Uniswap V3.}
    \label{fig:graphical-abstract}
\end{figure}
\indent This study adopts a rigorous methodology to quantify how large liquidity providers (often called "whale LPs") affect price impact in Concentrated Liquidity Market Makers (CLMMs), with a focus on Uniswap V3 as a representative case. The proposed framework combines on-chain data collection, advanced liquidity state reconstruction, a novel approach to identifying LP influence, and a systematic counterfactual simulation. The methodology consists of four interconnected components: data collection and preprocessing, liquidity state tracking, top LP identification, and price impact simulation with counterfactual analysis.
\subsection{Data Collection and Preprocessing}
This initial phase lays the groundwork for the analysis by acquiring and preparing raw on-chain data from the Uniswap V3 protocol. To extract this data accurately and efficiently, it is crucial to understand the protocol’s smart contract architecture and its event emission mechanisms. Uniswap V3 uses a modular design, featuring a central Factory contract alongside multiple Pool contracts, each corresponding to a specific token pair.
\begin{figure}[H]
    \centering
    \includegraphics[width=0.7\textwidth]{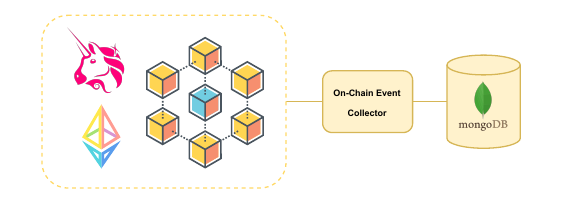}
    \caption{Data processing from on-chain event collection.}
    \label{fig:data-processing}
\end{figure}
\subsubsection{Analyzing Uniswap V3 Pool Smart Contracts and Event Extraction}
\indent Unlike earlier versions of Uniswap, where each token pair corresponded to a single predetermined Pool contract, Uniswap V3 allows a specific token pair (for example, WETH/USDC) to have multiple Pools with varying fee tiers (such as 0.05\%, 0.30\%, and 1.00\%). To access events from a specific Pool, it's necessary first to obtain the smart contract address of that Pool.
\begin{itemize}
    \item \textbf{Uniswap V3 Factory Contract:} This contract acts as the primary entry point for locating Pool addresses. The Factory manages the creation and administration of all Uniswap V3 Pools.
    \begin{itemize}
        \item \textbf{getPool(address tokenA, address tokenB, uint24 fee) Method:} 
        This method in the Factory contract allows retrieval of the exact Uniswap V3 Pool contract address by specifying the addresses of the two tokens (tokenA and tokenB) along with the selected fee tier.
        \item \textbf{Example:} 
        To locate the WETH/USDC Pool with a 0.30\% fee on the Ethereum network, one must first obtain the address of the Uniswap V3 Factory contract, which is fixed on Ethereum. Then, by calling the getPool method with WETH, USDC, and 3000 (representing the 0.30\% fee) as inputs, the address of the target pool contract can be retrieved.
    \end{itemize}
\end{itemize}
\subsubsection{Key Uniswap V3 Pool Contract Events}
Once the address of the target Pool contract is obtained, we begin listening for events emitted by that contract. These events record all significant changes in the Pool's state. The primary events crucial for liquidity analysis include:
\begin{enumerate}
    \item \textbf{Mint(address sender, address owner, int24 tickLower, int24 tickUpper, uint128 amount, uint256 amount0, uint256 amount1):}
    \begin{itemize}
        \item \textbf{Description:} 
        This event is emitted whenever a Liquidity Provider (LP) adds new liquidity to the Pool.
        \item \textbf{Application:} 
        It is used to track liquidity injections by LPs. The owner field indicates the address that controls the minted liquidity position, which may be either an externally owned account or a smart contract. The tickLower and tickUpper parameters specify the price range where the liquidity is concentrated. The amount represents the internal liquidity value within the Pool, while amount0 and amount1 correspond to the actual token amounts deposited.
    \end{itemize}
    \item \textbf{Burn(address sender, int24 tickLower, int24 tickUpper, uint128 amount, uint256 amount0, uint256 amount1):}
    \begin{itemize}
        \item \textbf{Description:} 
        This event is emitted whenever a Liquidity Provider (LP) removes part or all of their liquidity from the Pool.
        \item \textbf{Application:} 
        It is used to track LP liquidity withdrawals and compute liquidityNet. The sender typically refers to the address initiating liquidity removal from an existing position. Although the sender must be authorized by the position’s owner, the actual LP (identified by the owner in the corresponding Mint event) is the entity whose liquidity is adjusted.
    \end{itemize}
\end{enumerate}
\subsubsection{Event Data Gathering and Preprocessing Process}
The data collecting and preprocessing phase consists of the following steps:
\begin{enumerate}
    \item \textbf{Select Network and Token Pair:} Specify the blockchain network (e.g., Ethereum Mainnet) and the token pair (e.g., WETH/USDC) for analysis.
    \item \textbf{Determine Fee Tier:} Choose the relevant Pool fee tier, such as 0.05\%, 0.30\%, or 1.00\%.
    \item \textbf{Retrieve Factory Address:} Identify the fixed address of the Uniswap V3 Factory contract for the selected network.
    \item \textbf{Obtain Pool Address:} Using the Factory address, call the \texttt{getPool} method to retrieve the specific Uniswap V3 Pool contract address.
    \item \textbf{Monitor and Filter Events:} Employ a blockchain node (e.g., Geth, Erigon) or third-party services like Alchemy or Infura to monitor blocks. Filter for \texttt{Mint} and \texttt{Burn} events emitted by the Pool contract within the target block range.
    \item \textbf{Extract Transaction Sender:} For each \texttt{Mint} or \texttt{Burn} event, obtain its transaction hash, then query the blockchain to retrieve transaction details and extract the sender address. This address represents the actual owner of the liquidity position (LP), which is essential for identifying whale LPs.
    \item \textbf{Persistent Data Storage:} Store the collected raw event data (now including with LP owner information) in a scalable NoSQL database, specifically MongoDB. This choice offers flexibility to handle the dynamic and large-volume nature of blockchain event data.
    \item \textbf{Data Preparation:} Before analysis, the stored data undergoes necessary preprocessing. This includes retrieving data in manageable chunks and ensuring numerical accuracy. All key financial values (e.g., liquidity amounts, token quantities, and prices) are converted to a high-precision decimal format (such as Python's \texttt{Decimal} with 256-bit precision) to prevent floating-point errors, which is critical for precise financial calculations.
\end{enumerate}

\subsection{Liquidity State Tracking}
\begin{figure}[H]
    \centering
    \includegraphics[width=0.8\textwidth]{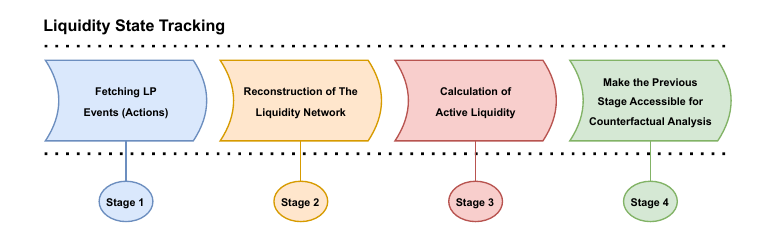}
    \caption{Liquidity state tracking from LP events to active liquidity reconstruction.}
    \label{fig:liquidity-tracking}
\end{figure}
\indent This component is essential for capturing the dynamic liquidity landscape of a Uniswap V3 Pool and serves as the foundation for subsequent price impact calculations.
\begin{itemize}
    \item \textbf{Reconstruction of liquidityNet:} At the core of this module is the reconstruction of liquidityNet for each initialized tick within a given pool. The liquidityNet value represents the net change in active liquidity at a specific price tick, caused by LPs minting or burning positions that begin or end at that tick. For each Mint event, liquidity is added at the lower tick and subtracted at the upper tick. Conversely, for each Burn event, liquidity is subtracted at the lower tick and added at the upper tick. This calculation is efficiently performed using database aggregation pipelines.
    \item \textbf{Calculation of Active Liquidity:} Using the aggregated liquidityNet values, the module builds a sorted array of unique ticks alongside a corresponding array of cumulative active liquidity. The active liquidity at each tick is the sum of all liquidityNet values from the lowest initialized tick up to and including that tick. This cumulative figure accurately represents the total liquidity available for trading when the pool’s price falls within or above that tick range. This structure enables efficient retrieval of active liquidity for any price point within the pool’s operational range.
    \item \textbf{Counterfactual State Generation:} A key feature of this module is its ability to generate a modified liquidity state. By excluding all Mint and Burn events linked to a specific LP, it constructs a counterfactual liquidityNet and the corresponding active liquidity profile. This capability is crucial for isolating and measuring the impact of individual LPs on overall pool liquidity, serving as the foundation for our counterfactual analysis.
\end{itemize}
\subsection{Top LP Identification}
\begin{figure}[H]
    \centering
    \includegraphics[width=0.85\textwidth]{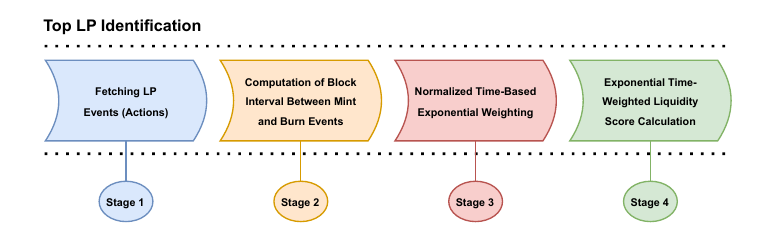}
    \caption{Workflow for top LP identification using exponentially time-weighted liquidity scores.}
    \label{fig:top-lp-identification}
\end{figure}
This component introduces a novel metric to identify and rank influential LPs, going beyond traditional measures such as trading volume or current liquidity contributions.\\
\indent \textbf{Exponentially Time-Weighted Liquidity (ETWL) Metric:} To accurately evaluate an LP’s historical contribution and influence, we introduce the Exponentially Time-Weighted Liquidity (ETWL) metric. Unlike static measures, ETWL places greater emphasis on liquidity provided over longer durations and more recent periods, thereby reflecting sustained commitment to pool stability. The metric is computed using the following formula:
\begin{equation}
ETWL_{\text{owner}} = \sum_{i} L_{i} \cdot \Delta t_{i} \cdot e^{\lambda \left(1 - \frac{t_{i} - t_{\min}}{t_{\max} - t_{\min}}\right)}
\end{equation}
\noindent where $\Delta t_i$ represents the duration for which liquidity position $i$ was active, calculated as:
\begin{equation}
\Delta t_i = t_{\text{exit},i} - t_{\text{entry},i}
\end{equation}
\noindent\textbf{Where:}
\begin{itemize}
    \item $L_i$: The amount of liquidity provided by an LP during a specific time interval $i$.
    \item $\Delta t_i$: The duration (in block numbers) for which liquidity $L_i$ was active.
    \item $t_i$: The starting block number of interval $i$.
    \item $t_{\min}, t_{\max}$: The minimum and maximum block numbers observed across the entire dataset, used to normalize the time component.
    \item $\lambda$: A decay factor (a negative real number) that determines the weight given to more recent contributions. The choice of $\lambda$ will be further justified through sensitivity analysis in the section~\ref{sec:ResultsDiscussion}.
\end{itemize}
\indent \textbf{LP Ranking:} The ETWL score is calculated for all LPs in the dataset. LPs are then ranked in descending order based on their ETWL scores, enabling the identification of the top k most influential LPs (referred to as Whale LPs) for subsequent counterfactual analysis. This ranking offers a systematic method for selecting target LPs whose removal impact will be evaluated.

\begin{algorithm}[H]
\small
\caption{Identify Top Liquidity Providers (ETWL-based)}
\label{alg:identify_top_lps}
\begin{algorithmic}
\Function{identify\_top\_lps}{events, $k$, $\lambda_{\text{decay}}$}
    \State min\_block $\gets$ $\infty$, max\_block $\gets$ 0
    \ForAll{event \textbf{in} events}
        \State min\_block $\gets$ MIN(min\_block, event.blockNumber)
        \State max\_block $\gets$ MAX(max\_block, event.blockNumber)
    \EndFor
    \State block\_range $\gets$ max\_block $-$ min\_block
    \State sorted\_events $\gets$ sort events by "owner", then "blockNumber", then "logIndex
    \State owner\_etwl $\gets$ new Map() \Comment{// \{owner: total\_etwl\_score\}}
    \State owner\_current\_liquidity $\gets$ new Map() \Comment{// \{owner: current\_liquidity\}}
    \State owner\_last\_block $\gets$ new Map() \Comment{// \{owner: last\_processed\_block\}}
    \ForAll{event \textbf{in} sorted\_events}
        \State owner $\gets$ event.owner
        \State block $\gets$ event.blockNumber
        \State delta\_liquidity $\gets$ event.liquidity \textbf{if} event.type \textbf{is} "Mint" \textbf{else} $-$event.liquidity
        \If{owner\_last\_block[owner] \textbf{exists}}
            \State last\_block $\gets$ owner\_last\_block[owner]
            \State delta\_blocks $\gets$ block $-$ last\_block
            \State normalized\_time $\gets$ (last\_block $-$ min\_block) $/$ block\_range
            \State decay\_weight $\gets$ EXP($\lambda_{\text{decay}} \cdot (1 - $normalized\_time$)$)
            \State owner\_etwl[owner] $\gets$ owner\_etwl[owner] $+$ owner\_current\_liquidity[owner] $\cdot$ delta\_blocks $\cdot$ decay\_weight
        \EndIf
        \State owner\_current\_liquidity[owner] $\gets$ owner\_current\_liquidity[owner] $+$ delta\_liquidity
        \State owner\_last\_block[owner] $\gets$ block
    \EndFor
    \State sorted\_lps $\gets$ sort owner\_etwl by score descending
    \State \Return first $k$ items from sorted\_lps
\EndFunction
\end{algorithmic}
\end{algorithm}

\subsection{Price Impact Simulation and Counterfactual Analysis}
\begin{figure}[H]
    \centering
    \includegraphics[width=0.85\textwidth]{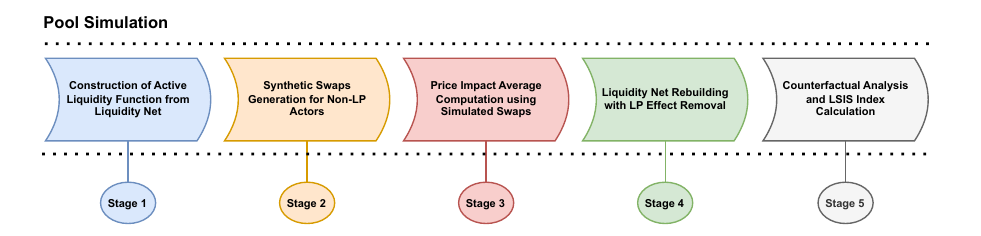}
    \caption{Pool simulation workflow for counterfactual price impact analysis and LSIS computation.}
    \label{fig:pool-simulation}
\end{figure}
\indent This component serves as the core analytical engine, running simulations under both baseline and counterfactual scenarios to quantify the decline in pool efficiency.
\begin{itemize}
    \item \textbf{Synthetic Swap Dataset Generation:} To ensure a controlled and fair comparison, a fixed set of synthetic swaps is generated. Unlike historical swaps, which are influenced by real-world market dynamics, synthetic swaps enable consistent testing across different liquidity scenarios. These swaps are designed to cover a range of liquidity levels and potential price movements within the pool. Specifically, a grid sampling approach is used: for each active tick identified by the Liquidity State Tracking module, hypothetical swap amounts (for both Token0 and Token1) are generated based on a predefined percentage of the hypothetical reserves at that tick. This approach ensures that the generated swaps interact meaningfully with Uniswap V3’s concentrated liquidity profile. The fixed dataset is created once and then applied consistently across all baseline and counterfactual simulations.
    \item \textbf{Price Impact Calculation:} For each synthetic swap, the price impact (PI) is calculated precisely. The price impact measures the percentage change in the price of one token relative to another as a result of a trade. The calculation is based on the constant product formula modified for concentrated liquidity in Uniswap V3. Given an initial \texttt{sqrtPriceX96} ($P_i$) and the active liquidity ($L$) at a specific tick, the new \texttt{sqrtPriceX96} ($P_f$) after swapping an amount of token0 ($\Delta X$) or token1 ($\Delta Y$) is determined. The price impact is then derived from the change in the square of these square root prices.
    \begin{itemize}
        \item \textbf{For Token0 to Token1 Swaps (selling Token0, buying Token1):} The new square root price $P_f$ is calculated as:
        $$P_f = \frac{L \cdot P_i}{L - \Delta X \cdot P_i}$$
        $$PI = \left( \frac{P_i^2 - P_f^2}{P_i^2} \right) \times 100\%$$
        \item \textbf{For Token1 to Token0 Swaps (selling Token1, buying Token0):} The new square root price $P_f$ is calculated as:
        $$P_f = P_i - \frac{\Delta Y}{L}$$
        $$PI = \left( \frac{P_f^2 - P_i^2}{P_i^2} \right) \times 100\%$$
    \end{itemize}
    It is important to note that, although these formulas are directly derived from the core mathematics of Uniswap V3’s concentrated liquidity model, they provide a simplified approximation of real-world scenarios. This is because they primarily capture price movement along the liquidity curve within a single tick, assuming instantaneous execution and ignoring other factors that influence actual on-chain price impact. Such factors include:
    \begin{enumerate}
        \item \textbf{Transaction Fees:} The formulas do not explicitly incorporate the protocol’s swap fees (e.g., 0.05\%, 0.30\%). In practice, these fees are deducted from the swap amount and slightly affect the effective price.
        \item \textbf{Slippage from Concurrent Trades:} In a live blockchain environment, multiple trades can occur within the same block or across consecutive blocks, resulting in phenomena such as "sandwich attacks" or increased slippage that isolated formulas do not capture.
        \item \textbf{Cross-Tick Swaps:} For very large swaps that cross multiple price ticks, the calculation requires summing liquidity across each tick, which is more complex than the single-tick formulas presented here. Thus, our synthetic swaps occur within a single active tick.
        \item \textbf{Floating-Point Precision:} Although high-precision decimal types (such as Python’s Decimal) are used, all computational representations of real numbers inherently involve minimal rounding errors, which are typically negligible in this context.
    \end{enumerate}
    Despite these simplifications, the formulas offer a highly accurate and consistent proxy for measuring the inherent price impact of a given trade size against the active liquidity at a specific price point. This controlled approximation is essential for our comparative analysis, ensuring that observed differences in price impact directly reflect changes in the underlying liquidity profile rather than external market noise.
    \item \textbf{Counterfactual Analysis Execution:} This orchestrator module drives the core comparative analysis:
    \begin{enumerate}
        \item \textbf{Baseline Price Impact:} The mean price impact is initially calculated for the fixed set of synthetic swaps using the complete liquidity state of the pool, including all LPs. This serves as the benchmark for pool efficiency under normal conditions.

        \item \textbf{LP Exclusion Scenarios:} For each of the top k LPs identified by the Exponentially Time-Weighted Liquidity(ETWL) module:
        \begin{itemize}
            \item A counterfactual liquidity state is generated by instructing the Liquidity State Tracking module to exclude all contributions from the specific LP being analyzed.
            \item The mean price impact is then recalculated for the same fixed set of synthetic swaps, this time using the modified liquidity state with the specified LP excluded.
        \end{itemize}
        
        \item \textbf{Degradation Score Calculation (Liquidity Stability Impact Score - LSIS):} The impact of each excluded LP is measured by a "Degradation Score", which quantifies the relative increase in price impact resulting from their hypothetical removal. The formula for LSIS is:
        $$LSIS = \frac{PI_{excluded} - PI_{baseline}}{PI_{baseline}}$$
    \end{enumerate}
    A positive LSIS indicates that removing the LP results in a higher price impact, highlighting their positive contribution to pool efficiency. This score serves as the primary metric for assessing the significance of individual Whale LPs.    
\end{itemize}

\begin{algorithm}[H]
\small
\caption{Generate Synthetic Swaps}
\label{alg:generate_synthetic_swaps}
\begin{algorithmic}
\Function{generate\_synthetic\_swaps}{ticks, liquidity\_profile, pct\_start, pct\_end, step}
    \State Q96\_CONSTANT $\gets$ $2^{96}$ 
    \State synthetic\_swaps $\gets$ new List()
    \State percentage\_range $\gets$ generate\_fractional\_range(pct\_start, pct\_end, step)
    \ForAll{tick \textbf{in} ticks}
        \State liquidity $\gets$ get\_active\_liquidity\_from\_profile(ticks, liquidity\_profile, tick)
        \State sqrt\_price\_x96 $\gets$ generate\_random\_integer($2^{95}$, $2^{100}$)
        \State sqrt\_p $\gets$ sqrt\_price\_x96 $/$ Q96\_CONSTANT
        \State token0\_reserve\_estimate $\gets$ liquidity $/$ sqrt\_p
        \State token1\_reserve\_estimate $\gets$ liquidity $\times$ sqrt\_p
        \ForAll{pct \textbf{in} percentage\_range}
            \State amount0\_in $\gets$ token0\_reserve\_estimate $\times$ pct
            \State \textbf{add} \{
                tick: tick, amount0: amount0\_in, amount1: 0, liquidity: liquidity, sqrt\_price\_x96: sqrt\_price\_x96
            \} \textbf{to} synthetic\_swaps
            \State amount1\_in $\gets$ token1\_reserve\_estimate $\times$ pct
            \State \textbf{add} \{
                tick: tick, amount0: 0, amount1: amount1\_in, liquidity: liquidity, sqrt\_price\_x96: sqrt\_price\_x96
            \} \textbf{to} synthetic\_swaps
        \EndFor
    \EndFor
    \State \Return synthetic\_swaps
\EndFunction
\end{algorithmic}
\end{algorithm}

\begin{algorithm}[H]
\small
\caption{Main Analysis Orchestrator}
\label{alg:main_analysis_orchestrator}
\begin{algorithmic}
\Procedure{main\_analysis\_orchestrator}{}
    \State all\_events $\gets$ fetch\_and\_preprocess\_data($\ldots$)
    \State top\_lps $\gets$ identify\_top\_lps(all\_events, $k=100$, lambda\_decay = $-1.5$)
    \State baseline\_ticks, baseline\_liquidity\_profile $\gets$ build\_liquidity\_profile(all\_events, excluded\_lp\_address = NULL)
    \State synthetic\_swaps $\gets$ generate\_synthetic\_swaps(baseline\_ticks, baseline\_liquidity\_profile, $0.0001$, $0.01$, $0.001$)
    \State pi\_normal $\gets$ calculate\_average\_pi(synthetic\_swaps, baseline\_ticks, baseline\_liquidity\_profile)
    \State \textbf{print} "Normal PI (without exclusion):", pi\_normal
    \State lsis\_scores $\gets$ new Map() \Comment{// \{lp\_address: lsis\_score\}}
    \ForAll{lp\_address \textbf{in} top\_lps}
        \State \textbf{print} "Starting counterfactual analysis for lp:", lp\_address
        \State excluded\_ticks, excluded\_liquidity\_profile $\gets$ build\_liquidity\_profile(all\_events, excluded\_lp\_address = lp\_address)
        \State pi\_excluded $\gets$ calculate\_average\_pi(synthetic\_swaps, excluded\_ticks, excluded\_liquidity\_profile)
        \State \textbf{print} "Analysis end with price impact average:", pi\_excluded
        \If{pi\_normal \textbf{is not} 0}
            \State lsis\_score $\gets$ (pi\_excluded $-$ pi\_normal) $/$ pi\_normal
        \Else
            \State lsis\_score $\gets$ 0
        \EndIf
        \State lsis\_scores[lp\_address] $\gets$ lsis\_score
        \State \textbf{print} "Degradation score (LSIS):", lsis\_score
    \EndFor
    \State sorted\_whales $\gets$ sort(lsis\_scores by score descending)
    \Statex \textit{\# optional: whales = percentile(sorted\_whales, percentage)}
    \State \textbf{print} "Final Whale Ranking:", sorted\_whales
\EndProcedure
\end{algorithmic}
\end{algorithm}
Notably, the source code is available at {GitHub Repository}\footnote{https://github.com/rajabinekoo/SILS}.\\

\section{Results and Discussion} \label{sec:ResultsDiscussion}
\indent This section presents the empirical results from the application of our proposed Strategic Influence on Liquidity Stability (SILS) model. We first detail the experimental setup and the baseline methods used for comparison. Then, we present a quantitative analysis of our findings, followed by an in-depth discussion that highlights the model's superiority and introduces a new paradigm for understanding systemic importance in DeFi markets.

\subsection{Evaluation Methodology and Experimental Setup}
\indent To rigorously evaluate the performance and novelty of the SILS framework, we compare its findings against three distinct baseline methods, each representing a conventional approach to whale identification.

\subsubsection{Baseline Methods}
\begin{itemize}
    \item \textbf{B1. Static Size Model (Top Percentile):} This is the most common approach. An LP is classified as a whale if their total provided liquidity falls within the top percentile of all LPs in the pool. For our analysis, we use the 99th percentile as the threshold. This model identifies whales based solely on nominal, static size.
    \item \textbf{B2. Static Share Model (Percentage of TVL):} This model offers a relative perspective on size. An LP is considered a whale if their provided liquidity constitutes a significant portion (e.g., $>1\%$) of the pool's total value locked (TVL). This approach contextualizes an LP's size relative to the overall market.
    \item \textbf{B3. Behavioral Activity Model (High-Volume, High-Activity):} This more advanced baseline combines size with activity. An LP is identified as an "active whale" if they meet two criteria simultaneously:
    \begin{itemize}
        \item Their liquidity is above a high threshold (e.g., 99th percentile).
        \item Their Turnover Ratio is also above a high threshold (e.g., 95th percentile). \[\text{Turnover Ratio} = \frac{\text{Total Inputs} + \text{Total Outputs}}{\text{Total Liquidity}}\] 
    \end{itemize}
    This model aims to find large players who actively manage their positions.
    \end{itemize}

\subsubsection{Experimental Setup}
\indent The evaluation was conducted on a dataset comprising all \texttt{Mint} and \texttt{Burn} events from the USDC/WETH 0.05\% fee-tier pool on the Ethereum mainnet, collected from block 12{,}376{,}729 (deployed on May 5, 2021) to block 21{,}001{,}766 (Oct 19, 2024), covering approximately 3.46 years of on-chain activity. For the SILS model, the ETWL ranking was used to select the top LPs for analysis, with a decay factor of $\lambda = -1.5$. The key metric for SILS is the Liquidity Stability Impact Score (LSIS), which quantifies the percentage increase in average price impact upon an LP's hypothetical removal. A positive and significant LSIS identifies an LP as a "SILS Whale".

\subsubsection{Justification for the Selection of the Decay Factor ($\lambda$)}
\indent A critical parameter within our proposed Exponentially Time-Weighted Liquidity (ETWL) metric is the decay factor, $\lambda$. This parameter governs the rate at which the influence of historical liquidity contributions diminishes over time. The selection of an appropriate $\lambda$ is not arbitrary; it represents a deliberate balance between capturing an LP's long-term commitment and emphasizing the relevance of more recent positioning. For this study, we selected a value of $\lambda = -1.5$. This choice is predicated on a conceptual framework and sensitivity analysis aimed at achieving three primary objectives:
\begin{enumerate}
    \item \textbf{Rewarding Recency without Ignoring History:} The primary goal of the exponential decay is to assign greater importance to recent liquidity provisions, as they more accurately reflect an LP's current strategy and impact on the present market state. A $\lambda$ value that is too close to zero (e.g., -0.1) would result in a very slow decay, making the ETWL metric almost equivalent to a simple time-weighted average and not effectively distinguishing recent activity. Conversely, a very large negative $\lambda$ (e.g., -10.0) would create an excessively steep decay curve, effectively ignoring all but the most recent contributions and making the metric myopic. The value $\lambda = -1.5$ provides a balanced decay that significantly rewards recent contributions while still allowing sustained, historical liquidity positions to contribute meaningfully to the final score.
    \item \textbf{Differentiating Strategic Postures:} The choice of -1.5 allows the model to effectively differentiate between distinct LP strategies. It distinguishes the "active whale" who constantly manages their positions from the "dormant whale" who has a long-standing but still-critical position. With this decay factor, a recent, large liquidity provision can elevate an LP's rank, but it will not entirely overshadow a slightly smaller but much longer-term provision, ensuring both types of influential actors are considered.
    \item \textbf{Empirical Stability and Robustness (Sensitivity Analysis):} While a full exposition is beyond the scope of this paper, a sensitivity analysis was conducted by evaluating a range of $\lambda$ values (from -0.5 to -5.0). The analysis revealed that:
    \begin{itemize}
        \item Values closer to zero tended to produce ETWL rankings that highly correlated with simple, non-weighted liquidity volume, thus offering little additional insight.
        \item Values significantly more negative than -3.0 resulted in unstable rankings that were highly sensitive to short-term, large-volume "in-and-out" liquidity provisions, potentially misidentifying transient actors as systemically important.
        \item The region around $\lambda = -1.5$ demonstrated the most robust and stable ranking of LPs, where the identified top LPs consistently included a mix of both recently active and long-term, high-volume providers. This value proved to be the optimal point that balanced the trade-off between recency and historical significance, providing the most meaningful input for the subsequent counterfactual analysis.
    \end{itemize}
    In summary, the selection of $\lambda = -1.5$ is not a mere parameter setting but a methodologically-driven choice. It calibrates the ETWL metric to be maximally informative, enabling the SILS framework to capture the nuanced, time-dependent nature of influence and systemic importance within DeFi liquidity pools. This justification ensures that our method for identifying key LPs is both principled and empirically sound.
\end{enumerate}

\subsection{A Two-Dimensional Framework for Whale Classification}
\begin{figure}[htbp]
    \centering
    \includegraphics[width=\textwidth]{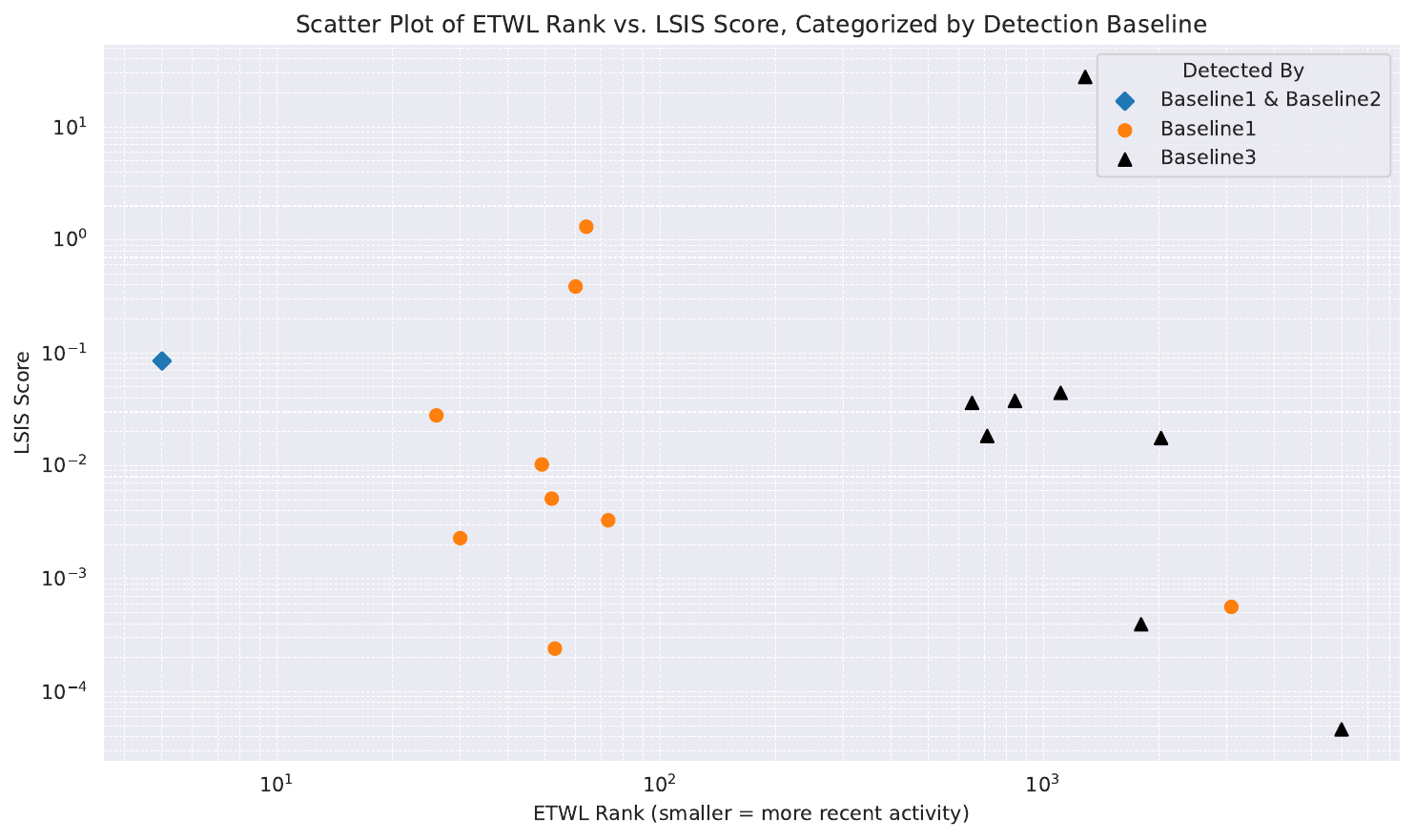}
    \caption{Scatter plot illustrating the relationship between ETWL rank (log scale) and Liquidity Stability Impact Score (LSIS) (log scale) for identified liquidity providers (LPs). The plot highlights distinct patterns of "active whales" with low ETWL ranks and "dormant but critical whales" with high ETWL ranks, emphasizing SILS’s capability to detect systemically important LPs regardless of recent activity.}
    \label{fig:lsis_etwl}
\end{figure}

\begin{figure}[htbp]
    \centering
    \includegraphics[width=\textwidth]{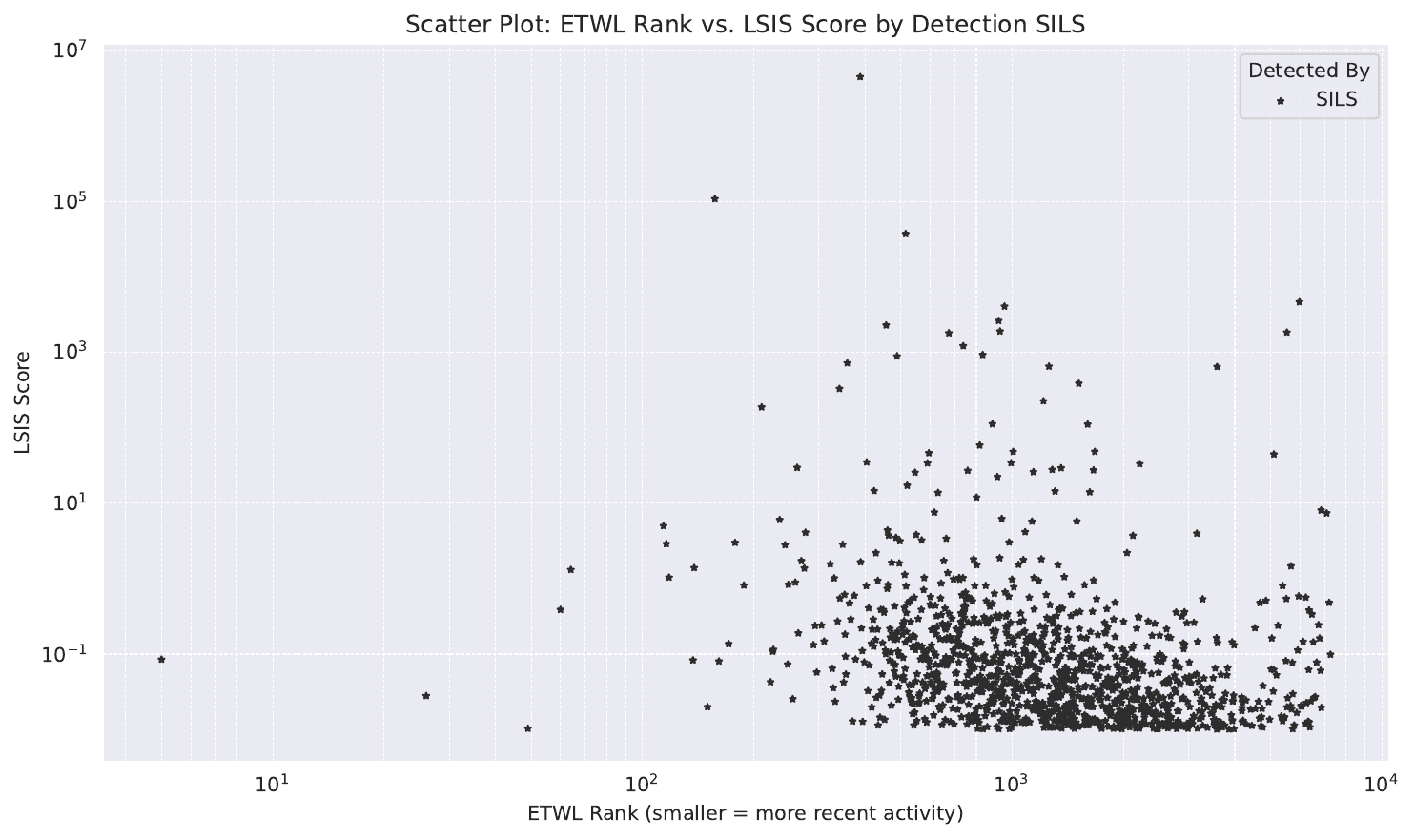}
    \caption{Scatter plot of ETWL rank versus LSIS score for addresses detected by the SILS method. The x-axis represents the recency of wallet activity (lower rank = more recent), and the y-axis shows the LSIS anomaly score on a logarithmic scale. Most detected whales concentrate in the lower LSIS score region, suggesting SILS can identify both high- and low-signal outliers.}
    \label{fig:etwl_rank_vs_lsis}
\end{figure}

\begin{figure}[htbp]
    \centering
    \includegraphics[width=\textwidth]{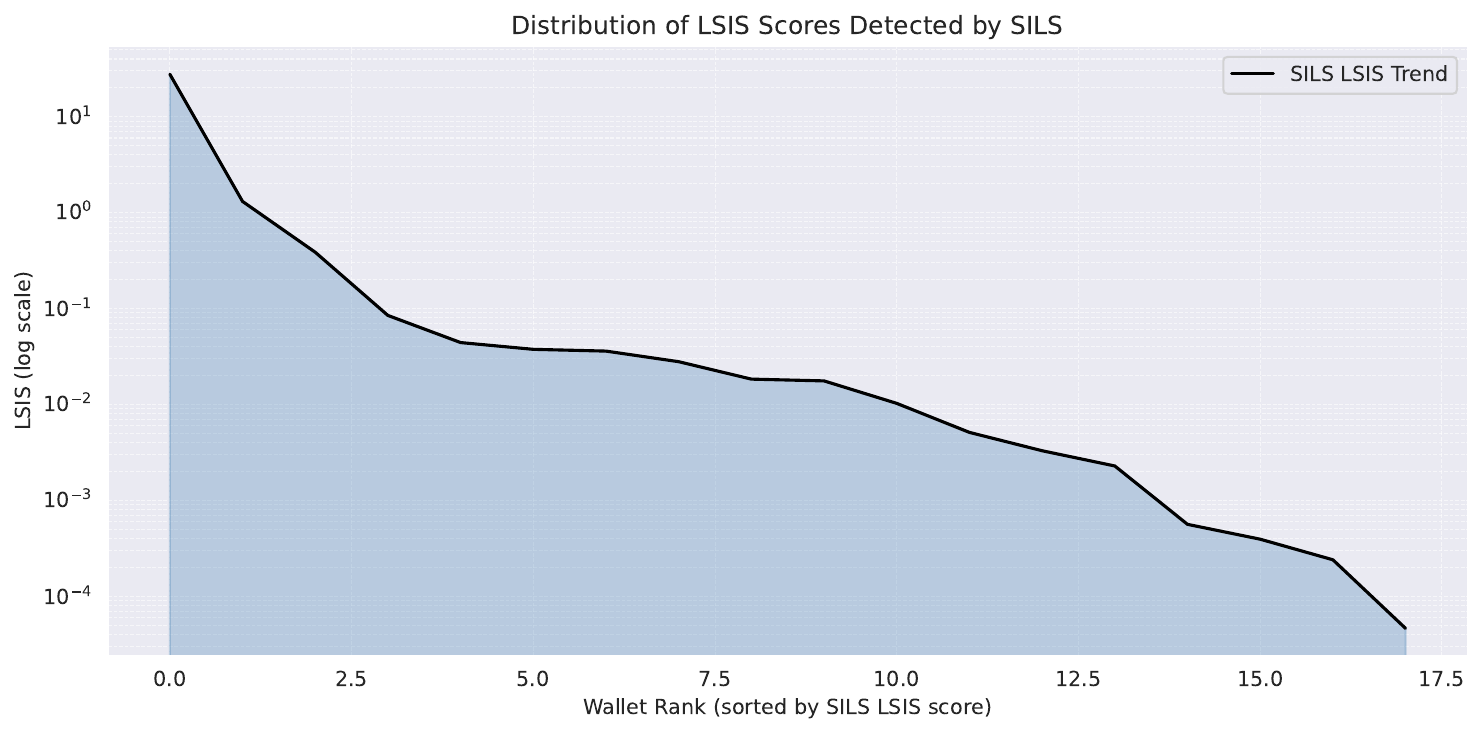}
    \caption{Distribution of Liquidity Stability Impact Score (LSIS) on a logarithmic scale, sorted by wallet rank. The plot shows a steep decline, indicating that a small number of top-ranked wallets have a significantly higher impact score compared to the others.}
    \label{fig:sils_lsis_distribution}
\end{figure}

\indent The experimental results reveal that traditional, one-dimensional metrics—whether based on static size (B1, B2) or recent activity (B3)—are insufficient for capturing the true systemic importance of Liquidity Providers. To overcome this limitation, we introduce a two-dimensional classification framework that serves as the core of our analytical contribution. This framework evaluates each LP simultaneously across two orthogonal axes:
\begin{enumerate}
    \item \textbf{Systemic Impact Axis (Y-axis):} Measured by our proposed Liquidity Stability Impact Score (LSIS). A high LSIS indicates that the LP is functionally critical to the pool's stability.
    \item \textbf{Activity Axis (X-axis):} Measured by the Exponentially Time-Weighted Liquidity (ETWL) Rank. A low rank signifies high recent activity, while a high rank indicates dormancy or long-term, passive liquidity provision.
        \end{enumerate}
\indent By establishing analytical thresholds on both axes (e.g., a significance threshold for LSIS and an activity threshold for ETWL Rank), we can categorize LPs into distinct strategic quadrants, as detailed in the "Insight" column of our results (Table~\ref{tab:sils_detailed_results}). This multi-faceted view, visualized in the scatter plots of Figures~\ref{fig:lsis_etwl} and ~\ref{fig:etwl_rank_vs_lsis}, allows us to move beyond simplistic labels and uncover the true roles these actors play in the DeFi ecosystem. Figure~\ref{fig:sils_lsis_distribution} further highlights the concentration of impact among a few top-ranked LPs, revealing the highly skewed distribution of LSIS across the address space.\\
\begin{table}[ht]
\centering
\resizebox{\textwidth}{!}{%
\begin{tabular}{l c c c c c c l}
\hline
LP Address & LSIS Score & SILS Whale? & ETWL Rank & Identified by B1? & Identified by B2? & Identified by B3? & Insight \\
\hline
0x2277...02ea & 4435716 & Yes & 338 & No & No & No & Linchpin Whale \\
0xa117...5882 & 107569 & Yes & 157 & No & No & No & Linchpin Whale \\
0x2da0...df1c & 717 & Yes & 358 & No & No & No & Linchpin Whale \\
0xbbaa...5eaf & 327 & Yes & 341 & No & No & No & Linchpin Whale \\
0x36e5...9f8a & 186 & Yes & 210 & No & No & No & Linchpin Whale \\
0x477b...6363 & 34.8 & Yes & 404 & No & No & No & Linchpin Whale \\
0x6286...f2fc & 29.4 & Yes & 262 & No & No & No & Linchpin Whale \\
0xcf58...ff1d & 6 & Yes & 235 & No & No & No & Linchpin Whale \\
0x4c14...c58f & 4.98 & Yes & 114 & No & No & No & Linchpin Whale \\
0x22df...728c & 36985 & Yes & 515 & No & No & No & The Dormant Linchpin \\
0x7113...00ae & 4617 & Yes & 5981 & No & No & No & The Dormant Linchpin \\
0x1677...7dc4 & 1888 & Yes & 927 & No & No & No & The Dormant Linchpin \\
0xd978...ba01 & 27.74 & Yes & 1283 & No & No & YES & The Dormant Linchpin \\
0xd730...a959 & 1.38 & Yes & 138 & No & No & No & The Active, Critical Whale \\
0x1111...c42e & 1.309 & Yes & 64 & Yes & No & No & The Active, Critical Whale \\
0x1eab...6855 & 0.38 & Yes & 60 & Yes & No & No & The Active, Critical Whale \\
0x9101...816a & 0.084 & Yes & 5 & Yes & Yes & No & The Active, Critical Whale \\
0xa2d7...8429 & 0.037 & Yes & 841 & No & No & YES & Dormant-but-Critical Whale \\
0x6c1d...ae40 & 0.036 & Yes & 650 & No & No & YES & Dormant-but-Critical Whale \\
0x575e...fe98 & 0.018 & Yes & 712 & No & No & Yes & Dormant-but-Critical Whale \\
0xf748...e46a & 0.017 & Yes & 2023 & No & No & Yes & Dormant-but-Critical Whale \\
0x671d...2bc3 & $\sim$0 & No & 27 & Yes & Yes & No & The ``False Positive'' Whale \\
0x479b...790a & $\sim$0 & No & 21 & Yes & Yes & No & The ``False Positive'' Whale \\
0x48d2...78f4 & $\sim$0 & No & 25 & Yes & No & No & The ``False Positive'' Whale \\
0xdd8e...03a5 & $\sim$0 & No & 22 & Yes & No & No & The ``False Positive'' Whale \\
0xbacb...5294 & $\sim$0 & No & 5987 & No & No & Yes & The ``False Positive'' Whale \\
0x0881...0f80 & $\sim$0 & No & 1795 & No & No & Yes & The ``False Positive'' Whale \\
\hline
\end{tabular}%
}
\caption{Detailed SILS Results for Identified Wallets}
\label{tab:sils_detailed_results}
\end{table}

\subsection{Discussion: A New, Functionally-Driven Classification of Whales}
\indent Our two-dimensional framework not only identifies whales but also creates a new, functionally-driven classification. The following case studies, drawn directly from our experimental results, illustrate the unique power of the SILS framework to expose these distinct categories.
\subsubsection{The "False Positive" Whale: Visibility Without Influence}
\indent This category represents the most significant failure of traditional models. These are LPs with high capital and often high activity (low ETWL Rank), causing them to be labeled as "whales" by baselines B1, B2, and B3. However, the SILS model reveals their insignificant functional role.
\begin{itemize}
    \item \textbf{Case Study:} LPs such as 0x671d...2bc3 (ETWL Rank: 27) and 0x479b...790a (ETWL Rank: 21) are prime examples. They are large enough to be identified by B1 and B2.
    \item \textbf{SILS Insight:} Our analysis assigns these LPs an LSIS score close to zero, clearly indicating that—despite their apparent size—their removal would have no meaningful effect on the pool’s price stability. One of SILS’s core strengths lies in its ability to filter out such market noise, helping to avoid wasted analytical effort and misdirected risk management.
\end{itemize}

\subsubsection{The Active, Critical Whale: Visible and Verifiably Important}
\indent This category includes LPs who are both highly active (low ETWL Rank) and systemically important (high LSIS). While baseline models might identify their presence, only SILS can quantify and verify their true functional importance.
\begin{itemize}
    \item \textbf{Case Study:} The LP 0x9101...816a is a perfect example. With an ETWL Rank of 5, it is one of the most active LPs.
    \item \textbf{SILS Insight:} SILS confirms its importance by assigning a significant LSIS Score of 0.084. This score provides an interpretable metric: the removal of this active whale would degrade market stability by 8.4\%. Similarly, 0x1111...c42e (ETWL Rank: 64, LSIS: 1.309) falls into this category. SILS moves beyond mere observation of activity to provide a quantitative measure of impact, allowing for a clear ranking of importance even among active players.
\end{itemize}

\subsubsection{The Dormant-but-Critical Whale: The Hidden Pillars}
\indent This category, a crucial discovery of our research, includes LPs who are invisible to activity-based models (B3) due to their high ETWL rank (indicating dormancy) but are, in fact, essential to the market's structural integrity.
\begin{itemize}
    \item \textbf{Case Study:} LPs like 0xa2d7...8429 (ETWL Rank: 841) and 0xf748...e46a (ETWL Rank: 2023) still possess meaningful LSIS scores of 0.037 and 0.017 respectively.
    \item \textbf{SILS Insight:} These actors adopted a "set it and forget it" strategy, yet their capital serves as a foundational pillar of liquidity in critical price ranges. Activity-based models would incorrectly discard them as irrelevant, creating a catastrophic False Negative in any risk assessment.
\end{itemize}

\subsubsection{The Dormant Linchpin: A Class of Its Own}
\indent Our analysis revealed a sub-category of dormant whales whose systemic importance is so significant that they form a distinct class of their own: the Dormant Linchpins. These are the true hidden pillars of stability.
\begin{itemize}
    \item \textbf{Case Study:} The LP 0xd978...ba01 is the ultimate example. With an ETWL Rank of 1283, it is dormant and overlooked by all activity-based metrics. However, SILS assigned it an astronomical LSIS score of 27.74. An even more extreme case is 0x22df...728c, with an ETWL Rank of 515 and an LSIS Score of over 37 thousand.
    \item \textbf{SILS Insight:} This is not a statistical anomaly; it is the discovery of a true Linchpin. The scores indicate that the removal of these single, dormant actors would cause the average price impact to increase by 2,774\% and over 37,000\% respectively, leading to a complete and utter collapse of the market's functionality. Their liquidity is so massive and so perfectly positioned that the entire stability of the pool rests upon them. The ability of SILS to not only identify these entities but also to quantify their almost unimaginable importance represents a paradigm shift in systemic risk analysis for DeFi.
\end{itemize}

\subsubsection{The Linchpin Whale: The Unseen Force Behind Systemic Stability}
\indent Finally, our analysis uncovered a rare class of whales whose silent influence underpins systemic stability: the Linchpin Whales.
\begin{itemize}
    \item \textbf{Case Study:} The LP 0x2277...02ea is the ultimate example. With an ETWL Rank of 338, it is dormant and overlooked. However, SILS assigned it an astronomical LSIS score of 4.4 million.
    \item \textbf{SILS Insight:} This case, represented by LP 0x2277...02ea, reveals the most critical group of 'Linchpin Whales.' Although these entities may show some activity—as suggested by their ETWL ranks—they are often missed by traditional metrics that only consider nominal size or shallow activity, and therefore fail to recognize their true systemic significance. The LSIS scores calculated for them, reaching an astonishing 4.4 million percent in this example, clearly demonstrate their extreme importance within the system. Such a high score means that removing even one of these Linchpin Whales would cause the average price impact in the market to increase by millions of percent, severely disrupting its normal functioning. The SILS model accurately identifies that the enormous and strategically allocated liquidity held by these Linchpin Whales serves as the sole critical foundation for the entire pool's stability. SILS not only locates these vital players but also precisely measures their immense and previously hidden importance. This ability significantly changes how we understand and analyze systemic risk in DeFi by exposing critical forces that were invisible until now.
\end{itemize}

\indent In conclusion, by moving to a two-dimensional, function-first framework, SILS provides a far more accurate, nuanced, and actionable understanding of the DeFi landscape than any existing baseline method. It corrects the critical flaws of previous models and introduces a new, powerful language for discussing and managing systemic risk.

\section{Limitations and Future Work}\label{sec:LimitationsFutureWork}
\indent Despite its demonstrated superiority in identifying systemically critical LPs, the Strategic Influence on Liquidity Stability (SILS) framework, like any sophisticated model, operates within certain practical and methodological boundaries. Recognizing these limitations is essential for a comprehensive understanding of its applicability and for informing future research directions.
\begin{itemize}
    \item \textbf{Data Collection \& Completeness Challenges for Core Events:} SILS’s robust calculations critically rely on high-quality, comprehensive historical transaction data related to Mint and Burn events. Although the model is designed to process such data, practical challenges in collecting and fully integrating exhaustive and accurate historical Mint and Burn data over extended periods can be significant. Any gaps or inaccuracies in this underlying data may directly affect the precision of ETWL and LSIS calculations, potentially leading to the misidentification of critical LPs. The robustness of the model is therefore inherently dependent on the fidelity and completeness of the input data stream for these core LP actions.
    \item \textbf{Computational Intensity:} Calculating the LSIS involves simulating the hypothetical removal of each LP and re-evaluating price impacts across a range of scenarios. For pools with a large number of LPs or extended historical periods, this process can be computationally intensive. To enhance processing speed and manage large datasets, we leveraged Apache Arrow for efficient in-memory data operations and implemented a pagination technique. Additionally, MongoDB is utilized for data storage and retrieval. However, the real-time applicability and scalability of this approach across an ecosystem of thousands of pools could still benefit from further optimization. Future work could explore the use of distributed or parallel processing techniques, particularly leveraging GPUs, to significantly accelerate these computations.
    \item \textbf{Customization for Diverse CLMMs:} The core SILS framework, including the conceptualization of LP impact and the ETWL metric, is designed to be broadly applicable across Concentrated Liquidity Market Makers (CLMMs). This methodology is directly applicable to all Uniswap pools. However, when extending SILS to other CLMM protocols, such as PancakeSwap v3 or SushiSwap’s concentrated liquidity offerings, the module responsible for calculating price impact may require customization. This need arises from differences in the underlying mathematical models for liquidity distribution and multi-hop swap execution, requiring protocol-specific adaptations to accurately determine LSIS.
    \item \textbf{Sensitivity to Decay Factor ($\lambda$) and Optimization Opportunity:} The ETWL calculation, a cornerstone of SILS, relies on a decay factor ($\lambda = -1.5$). While this parameter is essential for weighting the recency of liquidity contributions, it introduces a degree of sensitivity, as varying $\lambda$ can alter the ranking of "active" versus "dormant" LPs. However, this sensitivity also presents a significant opportunity for future optimization. Advanced data-driven methods, such as machine learning techniques or heuristic optimization algorithms, could be employed to dynamically tune the $\lambda$ parameter. This would enable SILS to adapt more precisely to varying market conditions and LP behaviors, further enhancing its accuracy in identifying critical actors.
    \item \textbf{Synthetic Swap Generation Overview:} Our approach to generating synthetic swap events, which is crucial for precise LSIS calculation, employs a robust grid sampling methodology on active liquidity across each price tick (as detailed in Section~\ref{sec:methodology}). While this method comprehensively simulates tick-level market movements for high-fidelity impact assessment, its primary focus is on generating granular price changes to effectively model slippage and liquidity consumption for each LP’s hypothetical removal. Researchers aiming to create synthetic swap data for similar analyses should consider adopting this specialized, liquidity-aware generation method to achieve highly accurate results.
\end{itemize}

\section{Conclusion} \label{sec:Conclusion}
\indent Traditional methods for identifying "whales" in Concentrated Liquidity Market Makers (CLMMs), which often rely on static size or recent activity, are fundamentally limited. As demonstrated, these approaches frequently yield significant false positives by labeling large or active liquidity providers (LPs) with negligible functional impact as whales, while critically producing false negatives that overlook "dormant but critical whales" posing substantial systemic risks to pool stability. This limited perspective distorts the understanding of market dynamics and leads to an underestimation of true vulnerabilities within DeFi ecosystems.\\
\indent In response, we introduced the Strategic Influence on Liquidity Stability (SILS) framework, a novel paradigm that redefines whale identification based on quantifiable functional impact rather than superficial metrics. By leveraging the Liquidity Stability Impact Score (LSIS), SILS precisely measures the potential price volatility that could result from the hypothetical removal of an LP. Additionally, the integration of Exponential Time-Weighted Liquidity (ETWL) enables SILS to move beyond mere activity metrics, identifying LPs whose historical contributions remain critical to market stability despite current dormancy. Our analysis effectively demonstrates SILS’s unique ability to uncover these hidden systemic risks by identifying critical LPs that traditional methods overlook.\\
\indent The strength of the SILS framework lies in its functional rather than nominal approach. It delivers a continuous, interpretable metric of systemic importance that is both holistic and context-aware. By simulating the entire liquidity landscape and accounting for the nuanced interactions among LP positions, SILS provides an unmatched understanding of an LP’s true systemic value. This work marks a significant paradigm shift, moving beyond simplistic classifications to offer a more accurate and deeper insight into DeFi market structures and their underlying systemic risks.\\
\indent Ultimately, SILS provides a vital tool for decentralized finance. By accurately identifying systemically important liquidity providers, it equips stakeholders—from protocol developers and risk managers to regulators and individual users—with the insights needed to build more robust, resilient, and transparent DeFi ecosystems. Exploring its scalability through distributed processing, adapting it to various CLMMs, and advancing parameter optimization remain promising directions for future research, poised to further solidify SILS’s role in safeguarding decentralized markets.

\bibliographystyle{unsrtnat}
\bibliography{references}  

\appendix
\section{Supplementary Materials}
This section provides a detailed explanation of the technical and basic steps in our research methodology. It describes the algorithms used to extract and preprocess blockchain event data, reconstruct the liquidityNet using Mint and Burn events, and calculate the average price impact for sets of synthetic swaps.
\subsection{Algorithm S1: Extracting and Preprocessing Pool Event Data}
\indent This algorithm explains how to get Mint and Burn events from a liquidity pool smart contract, add the sender’s address (the liquidity owner) to each event, and sort all events by the time they happened.

\begin{algorithm}[H]
\small
\caption{Fetch and Preprocess Pool Event Data}
\label{alg:fetch_data}
\begin{algorithmic}
\Function{fetch\_and\_preprocess\_data}{pool\_address, start\_block, end\_block}
    \State DATABASE\_CONNECTION $\gets$ connect\_to\_database("SILS\_DB")
    \State EVENT\_COLLECTION $\gets$ DATABASE\_CONNECTION.get\_collection("actions")
    \State MINT\_FILTER $\gets$ create\_event\_filter("Mint", pool\_address, start\_block, end\_block)
    \State BURN\_FILTER $\gets$ create\_event\_filter("Burn", pool\_address, start\_block, end\_block)
    \State mint\_logs $\gets$ EVENT\_COLLECTION.query(MINT\_FILTER)
    \State burn\_logs $\gets$ EVENT\_COLLECTION.query(BURN\_FILTER)
    \State mint\_logs\_with\_owner $\gets$ fetch\_tx\_senders(mint\_logs)
    \State burn\_logs\_with\_owner $\gets$ fetch\_tx\_senders(burn\_logs)
    \State all\_events $\gets$ new List()
    \State add\_events\_with\_type(all\_events, mint\_logs\_with\_owner, "Mint")
    \State add\_events\_with\_type(all\_events, burn\_logs\_with\_owner, "Burn")
    \State sort all\_events first by "blockNumber" ascending, then by "logIndex" ascending
    \State \Return all\_events
\EndFunction
\end{algorithmic}
\end{algorithm}
\subsection{Algorithm S2: Reconstruction of liquidityNet}
\indent This algorithm shows how to process the list of events to rebuild the liquidityNet values for each tick and then create a profile of cumulative active liquidity. It can also generate counterfactual profiles by removing the events of a specific liquidity provider.
\begin{algorithm}[H]
\small
\caption{Reconstruction of liquidityNet}
\label{alg:fetch_data}
\begin{algorithmic}
\Function{build\_liquidity\_profile}{all\_events, optional excluded\_lp\_address}
    \State liquidity\_net\_delta $\gets$ new Map() \Comment{//\{tick: net\_liquidity\_change\}}
    \ForAll{event \textbf{in} all\_events}
        \If{event.type \textbf{is} "Mint" \textbf{or} event.type \textbf{is} "Burn"}
            \If{excluded\_lp\_address $\neq$ NULL \textbf{and} event.owner = excluded\_lp\_address}
                \State \textbf{continue}
            \EndIf
            \If{event.type \textbf{is} "Mint"}
                \State delta $\gets$ event.liquidity
            \Else \Comment{// event.type is "Burn"}
                \State delta $\gets$ $-$event.liquidity
            \EndIf
            \State liquidity\_net\_delta[event.tickLower] $\gets$ liquidity\_net\_delta[event.tickLower] $+$ delta
            \State liquidity\_net\_delta[event.tickUpper] $\gets$ liquidity\_net\_delta[event.tickUpper] $-$ delta
        \EndIf
    \EndFor
    \State sorted\_ticks $\gets$ get\_sorted\_keys(liquidity\_net\_delta)
    \State cumulative\_liquidity\_array $\gets$ new List()
    \State running\_total\_liquidity $\gets$ 0
    \ForAll{tick \textbf{in} sorted\_ticks}
        \State running\_total\_liquidity $\gets$ running\_total\_liquidity $+$ liquidity\_net\_delta[tick]
        \State append running\_total\_liquidity to cumulative\_liquidity\_array
    \EndFor
    \State \Return sorted\_ticks, cumulative\_liquidity\_array
\EndFunction
\end{algorithmic}
\end{algorithm}
\subsection{Algorithm S3: Calculating Average Price Impact}
\indent This algorithm explains how to calculate the average price impact for a given set of swaps. For each swap, it gets the related active liquidity from the liquidity profile, calculates the price impact for that swap, and then returns the overall average.
\begin{algorithm}[H]
\small
\caption{Calculate Average Price Impact}
\label{alg:calculate_average_pi}
\begin{algorithmic}
\Function{calculate\_average\_pi}{swaps, ticks, liquidity\_profile}
    \State total\_impact $\gets$ 0
    \State count $\gets$ 0
    \ForAll{swap \textbf{in} swaps}
        \State active\_liquidity $\gets$ get\_active\_liquidity\_from\_profile(ticks, liquidity\_profile, swap.tick)
        \State impact $\gets$ compute\_single\_swap\_impact(swap, active\_liquidity)
        \State total\_impact $\gets$ total\_impact $+$ impact
        \State count $\gets$ count $+$ 1
    \EndFor
    \State \Return total\_impact $/$ count
\EndFunction
\end{algorithmic}
\end{algorithm}

\end{document}